\documentclass{article}
\usepackage{spconf,amsmath,graphicx}
\usepackage{algorithmic}
\usepackage{algorithm}
\usepackage{hyperref}

\title{Working Alliance Transformer for Psychotherapy Dialogue Classification}
\name{Baihan Lin$^1$, Guillermo Cecchi$^2$, Djallel Bouneffouf$^2$}
\address{
  $^1$Columbia University\\
  $^2$IBM Thomas J. Watson Research Center\\
{baihan.lin@columbia.edu, gcecchi@us.ibm.com, djallel.bouneffouf@ibm.com}}
\begin{document}
%
\maketitle
\begin{abstract}

As a predictive measure of the treatment outcome in psychotherapy, the working alliance measures the agreement of the patient and the therapist in terms of their bond, task and goal. Long been a clinical quantity estimated by the patients' and therapists' self-evaluative reports, we believe that the working alliance can be better characterized using natural language processing technique directly in the dialogue transcribed in each therapy session. In this work, we propose the Working Alliance Transformer (WAT), a Transformer-based classification model that has a psychological state encoder which infers the working alliance scores by projecting the embedding of the dialogues turns onto the embedding space of the clinical inventory for working alliance. We evaluate our method in a real-world dataset with over 950 therapy sessions with anxiety, depression, schizophrenia and suicidal patients and demonstrate an empirical advantage of using information about the therapeutic states in this sequence classification task of psychotherapy dialogues. \footnote{Reproducible codes at
\href{https://github.com/doerlbh/PsychiatryNLP}{\underline{https://github.com/doerlbh/PsychiatryNLP}}.}

\end{abstract}

\section{Introduction}

The working alliance between the therapist and the patient is an important measure of the clinical outcome and a qualitative predictor of therapeutic effectiveness in psychotherapy \cite{Wampold2015,Bordin79}. The alliance entails a number of cognitive and emotional aspects of the interaction between these two agents, such as their shared understanding of the objectives to be attained and the tasks to be completed, as well as the bond, trust, and respect that will develop during the course of the therapy. While traditional methods to quantify the alliance depend on self-evaluative reports with point-scales valuation by patients and therapists about whole sessions \cite{horvath1981exploratory}, the digital era of mental health can enable new research fronts utilizing real-time transcripts of the dialogues between the patients and therapists. By analyzing the psychotherapy dialogues, we are interested in studying the usage of natural language processing technique to extract out turn-level features of the working alliance and see if it can help better inform us of the clinical condition of the patient.


Here we present the Working Alliance Transformer (WAT), a transformer-based classification model to classify the psychotherapy sessions into different psychiatric conditions. Our methods consists of a psychological state encoder that quantifies the degree of patient-therapist alliance by projecting each turn in a therapeutic session onto the representation of clinically established working alliance inventories, using language modeling to encode both turns and inventories, which was originally proposed in \cite{lin2022deep} as an analytical tool. This allows us not only to quantify the overall degree of alliance but also to identify granular patterns its dynamics over shorter and longer time scales. We collated and preprocessed the Counseling and Psychotherapy Transcripts from Alex Street the publisher company \footnote{https://alexanderstreet.com/products/counseling-and-psychotherapy-transcripts-series}, which consists of transcribed recordings of over 950 therapy sessions between multiple anonymized therapists and patients that belong to four types of psychiatric conditions: anxiety, depression, schizophrenia and suicidal. This multi-part collection includes speech-translated transcripts of the recordings from real therapy sessions, 40,000 pages of client narratives, and 25,000 pages of reference works.
On this dataset, we evaluate quantitatively the effectiveness of this inference method in improving the classification or diagnosis capability of deep learning models to predict psychiatric conditions from therapy transcripts.
Lastly, we discuss how our approach may be used as a companion tool to provide feedback to the therapist and to augment learning opportunities for training therapists.



 \section{Methods}

\begin{figure*}[tb]
\centering
    \includegraphics[width=\linewidth]{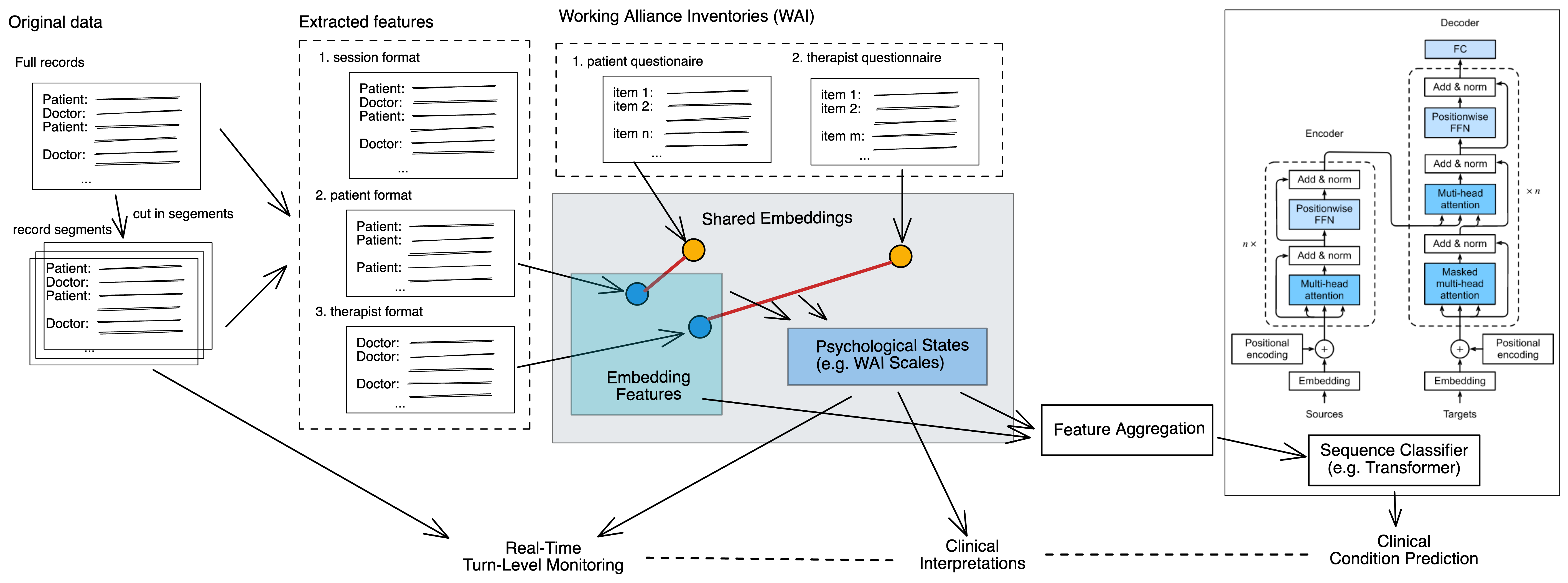}
\caption{Architecture of working alliance transformer for psychiatric condition classification using the psychological state encoder from working alliance
}\label{fig:classifier}
\end{figure*}

We describe our pipeline in Figure \ref{fig:classifier}. Given the transcripts of a therapy session and the medical records of the patient. The dialogue are separated into pairs of turns as the timestamps. We can either choose to only use the turns by the patients, or by the therapists, or use both, as a paired input. Empirically, the patients' turns are usually more narrative, as they are describing themselves, while the therapists' turns are usually more declarative, as they are usually confirming the patients, or leading the conversations to a certain topic.

On the other hand, we have access to the clinical inventory of the working alliance, the Working Alliance Inventory (WAI). The modern WAI consists of 36 questions or statements in a self-report questionnaire which measures the therapeutic bond, task agreement, and goal agreement \cite{horvath1981exploratory,tracey1989factor,martin2000relation}, where the 
the rater (i.e. the patient or the therapist) is asked to rate each statement on a 7-point scale (1=never, 7=always)\cite{martin2000relation}. This inventory is disorder-agnoistic, meaning that it measures the alliance factors across all types of therapies, and provides a record of the mapping from the alliance measurement and the corresponding cognitive constructs underlying the measurement under a unified theory of therapeutic change \cite{horvath1994working}.

\begin{algorithm}[tb]
 \caption{Working Alliance Transformer (WAT)}
 \label{alg:waa}
 \begin{algorithmic}[1]
 \STATE {\bfseries } \textbf{Input}: a session with $T$ turns
 \STATE {\bfseries } \textbf{Output}: a label for psychiatric condition
 \STATE {\bfseries }\textbf{for} i = 1,2,$\cdots$, T \textbf{do}
\STATE {\bfseries } \quad Automatically transcribe dialogue turn pairs  $(S^p_i,S^t_i)$
\STATE {\bfseries }\quad \textbf{for} $(I^p_j, I^t_j) \in$ inventories $(I^p, I^t)$ \textbf{do} \\
\STATE {\bfseries }\quad \quad Score $W^{p_i}_{j}$ = similarity($Emb({I^p_j}), Emb(S^p_i)$) \\
\STATE {\bfseries }\quad \quad Score $W^{t_i}_{j}$ = similarity($Emb({I^t_j}), Emb(S^t_i)$) \\
\STATE {\bfseries } \quad \textbf{end for}
\STATE {\bfseries }\quad Patient feature $x_c = concat(Emb(S^t_i), W^{t_i})$ 
\STATE {\bfseries }\quad Therapist feature $x_t = concat(Emb(S^t_i), W^{t_i})$ 
\STATE {\bfseries }\quad  Full feature $x = concat(x_t, x_c)$ 
\STATE {\bfseries }\quad Aggregated feature $X.append(x)$
\STATE {\bfseries } \textbf{end for}
\STATE {\bfseries } obtain prediction $y=Transformer(X)$ 
 \end{algorithmic}
\end{algorithm}

The inference goal is to compute a score that characterizes the working alliance given the clinical inventory, with for instance, a feature vector of 36 dimension that correspond to the 36 alliance measure of interests in the inventory. 
After computing the information regarding the predicted clinical outcome with our inferred working alliance scores, this feature vector highlights a bias towards what the clinicians would care about in the psychotherapy given the metrics provided by the working alliance inventory. We would then able to further use this information to potentially inform us of the psychiatric condition of a given patient. As such, we propose the Working Alliance Transformer (WAT), a classification model that utilizes an inference module that informs the downstream classifier where the current state is with respect to the therapeutic trajectory or landscape in the psychotherapy treatment of this patient. Is this patients approaching a breakthrough? Or is he or she susceptible to a rupture of trust? And these therapeutic information about working alliance can vary across clinical conditions, and as a result, potentially beneficial to the diagnosis and monitoring of the psychiatric disorders. 

Algorithm \ref{alg:waa} outlines the classification process. During the session, the dialogue between the patient and therapist are transcribed into pairs of turns.
We denote the patient turn as $S^p_i$ followed by the therapist turn $S^t_i$, as a dialogue pair. Similarly, the inventories of working alliance questionnaires come in pairs ($I^p$ for the patient, and  $I^t$ for the therapist, each with 36 statements). We compute the distributed representations of both the dialogue turns and the inventories with the sentence embeddings. The working alliance scores can then be computed as the cosine similarity between the embedding vectors of the turn and its corresponding inventory vectors. Following \cite{lin2022deep,lin2022unsupervised}, we use SentenceBERT \cite{reimers2019sentence} and Doc2Vec embedding \cite{le2014distributed} as our sentence embeddings for the working alliance inference.
With that, for each turn (either by patient or by therapist), we obtain a 36-dimension working alliance score.
For the classification, we concatenate the 36-dimension working alliance scores computed from the current turn in the dialogue, along with the sentence embedding of the current turn, as our feature vector to fed into our Transformer sequence classifier. 





The analytical features enabled by the working alliance inference are not only useful for the classification we investigate in this study but also other downstream tasks, such as predictive modeling and real-time analytics.
In our case, the turns in a dialogue or monologue are fed into the sentence embedding sequentially as individual entries. And then, given the sentence embedding, we feed them each into the psychological state encoder that infer the psychological or therapeutic state of the dialogue at this turn. The encoder will generate a vector that characterizes the state, such as the 36-dimension working alliance scores, corresponding to the 36 working alliance inventory items. Then, the model aggregate both the sentence embedding feature vector and the psychological state vector. 
Since we feed our input sentence by sentence (or turn by turn), we have a sequence of combined feature vector, which is then fed into a sequence classifier. We use the transformer \cite{vaswani2017attention} as our classifier for its effectiveness in various sequence-based learning tasks, and potential interpretability from its attention weights. The output of this classification model is the predicted clinical condition of this sequence. The sequence of turns we feed to generate a label is typically either the entirety or a segment of a session of psychotherapy transcript.




\section{Results}

\begin{table*}[t]
      \caption{Classification accuracy (\%) of psychotherapy sessions
      }
      \label{tab:classification} 
      \centering
      \resizebox{\linewidth}{!}{
 \begin{tabular}{ l | c | c | c | c | c | c }
  & \multicolumn{3}{c}{SentenceBERT} \vline & \multicolumn{3}{c}{Doc2Vec} \\
  & Patient turns & Therapist turns & Both turns & Patient turns & Therapist turns & Both turns \\ \hline
WAT (working alliance embedding) & \textbf{27.6} & \textbf{27.0} & \textbf{26.0} & \textbf{34.1} & 25.7 & \textbf{31.9} \\
WAT (working alliance score) & 26.1 & 23.4 & 25.5 & 28.9 & 23.7 & \textbf{31.9} \\
Embedding Transformer & 24.8 & 24.0 & 25.5 & 31.8 & \textbf{26.2} & 29.9 \\ \hline
WA-LSTM  (working alliance embedding) & \textbf{35.0} & \textbf{36.9} & \textbf{23.3} & \textbf{46.0} & 27.7 & 29.6 \\
WA-LSTM (working alliance score) & 24.5 & 34.2 & 22.6 & 30.2 & 24.7 (F) & \textbf{43.4} \\
Embedding LSTM & 23.0 & 36.0 & 22.9 & 44.3 & \textbf{31.1} & 31.1 \\ \hline
WA-RNN  (working alliance embedding) & 22.8 & \textbf{30.6} & 26.8 & 23.0 (F) & 24.9 & 19.1 \\
WA-RNN (working alliance score) & \textbf{30.5} & 28.0 (F) & 25.6 (F) & 24.0 (F) & 22.9 & \textbf{32.6} \\
Embedding RNN & 25.3 & 27.5 & \textbf{29.0} & \textbf{33.8} & \textbf{29.0} & 26.2 \\ \hline
\end{tabular}
 }
\end{table*}

In this section, we present the results of the session classification of psychotherapy dialogues into four clinical labels. 

\subsubsection{Experimental setting.} 

The psychotherapy dataset we evaluate is highly imbalanced across the four clinical conditions (495 anxiety sessions, 373 depression sessions, 71 schizophrenia sessions, and 12 suicidal sessions). If we directly train our models on this dataset, the classifier is likely to be highly biased towards the majority class. To correct for this imbalanceness issue, we are using the sampling technique. Instead of going through the entire training data in epochs, we train the models in sampling iterations. In each iteration we randomly choose a class and then randomly sample one session from the class pool. Before we sample the sessions, we split the dataset into 20/80 as our test set and training set. Then during the training or the test phase, we perform the sampling technique for each iteration only in the fully separated training and test sets. Then, for each sampled session, we feed into the classification model the first 50 dialogue turns of our transcript, turn by turn, and the sequence classifier will output a label predicting which psychiatric condition this session belongs to.

\subsubsection{Model architecture.} 

Following the model architecture introduce in the Method section and Figure \ref{fig:classifier}, we evaluate three classifier backbones. The first one is the classical transformer model. For the multi-head attention module, we set the number of heads to be 4 and the dimension of the hidden layer to be 64. The dropout rates for the positional encoding layer and the transformer blocks are both set to be 0.5. The second sequence classifier is a single-layer Long Short-Term Memory (LSTM) network \cite{hochreiter1997long} with 64 neurons. The third sequence classifier is a single-layer Recurrent Neural Network (RNN) with 64 neurons.

\subsubsection{Ablation and baseline models.} 

For each of the three classifiers, we compare three types of features as the input we feed into the sequence classifier component. The first one, the working alliance embedding, is the concatenated feature vector of both the sentence embedding vector and the psychological state vector (which in our case, is the 36-dimension inferred working alliance scores). The second type of feature, the working alliance score, is an ablation model which only uses the state vector (the working alliance score vector). The third type of feature, the embedding, is the baseline which only uses the sentence embedding vector directly. In other words, The working alliance score introduces the bias for WAI. The sentence embedding doesn't. The working alliance embedding is the feature that combines both with concatenation. And since we have two sentence embeddings to choose from (the sentence BERT and Doc2Vec), they each have 9 models in the evaluation pool. Other than the classifier types (Transformer, LSTM or RNN), the embedding types (SentenceBERT or Doc2Vec) and the feature types (working alliance embedding, working alliance scores, or simply sentence embedding), we also compared using only the dialogue turns from the patients, from the therapists, and from both the patients and the therapists. In the case where we use the turns from both the patients and the therapists, we consider them as a pair, and concatenate them together as a combined feature. This is as opposed to treating them as subsequent sequences, because we believe that the therapist's response are loosely semantic labels for the patient's statements, and thus, serve different semantic contexts that should be considered side by side, instead of sequentially, which would assume a homogeneity between time steps.

\subsubsection{Training procedure.} 

For all 18 models, we train them for over 50,000 iterations using the stochastic gradient descent with a learning rate of 0.001 and a momentum of 0.9. Since the training set is relatively small for our neural network models, we observe some of the models exhibit overfitting at early stages before we finish the training. As a result, we report the performance of their checkpoints where they converge and have a plateau performance. Then in the testing phase, we randomly sample 1,000 samples with equal probabilities in four classes.

\subsubsection{Empirical results.} 

We report the classification accuracy as our evaluation metrics. Since we have four classes, and the evaluation is corrected for imbalanceness with the sampling technique.
Table \ref{tab:classification} summarizes the results and the confusion matrices are reported in the supplementary materials. 

Overall, we observe a benefit of using the working alliance embedding as our features in Transformer and LSTM-based model architectures. Among all the models, the WA-LSTM model with working alliance embedding using only the patient turns obtains the best classification result (46\%), followed by the WA-LSTM model using only the working alliance score using both turns from the patients and therapists (43.4\%). This suggest the advantage of taking into account the predicted clinical outcomes in characterizing these sessions given their clinical conditions. We also notice that the inference of the therapeutic working alliance with Doc2Vec appears to be more beneficial in modeling the patient turns than the therapist turns, while the working alliance inference using SentenceBERT appears to be advantageous in both the therapist and patient features.

During training, we observe that among the three sequence classifier variants, the vanilla RNNs sometimes fail (which we denote ``F'') to learn due to exploding gradients over the long time steps (over 100 turns in each session). As a result, their prediction are at the chance level and based on their confusion matrices, they only trivially select the first class label. The LSTM networks are more stable when dealing with these long time series, but we do also observe one failure case when it is trained on the working alliance score of the therapists' turns as its features. 

Comparing the three sequential learners, the Transformer, due to the additional attention mechanism, yields a more stable learning phase.
When using the SentenceBERT as its embedding, we observe a modest benefit when training on only the patient turns, which might suggest an interference of features between the therapists' and patients' working alliance information. The Transformers using the working alliance embedding, i.e. both the sentence embedding and their therapeutic states (i.e. the inferred working alliance score vector) are the best performing ones. When using the Doc2Vec as our sentence embedding, the best performing models are both the Transformers using some of the working alliance information from our inference module as the features. These preliminary results suggest that the inferred scores of the therapeutic or psychological state can be potentially useful in downstream tasks, such as diagnosing the clinical conditions. Although not a main focus in this work, future work would include a more systematic investigation of such downstream tasks, as well as utilizing the attention mechanism of the transformer blocks for interpretations.

\section{Conclusions}

In this work, we present a Transformer-based classification model that characterizes the sequence of therapeutic states as beneficial feature to improve the classification of psychological dialogues into different psychiatric conditions. It combines the domain expertise from clinically validated psychiatry inventories with the distributed deep representations of language modeling provide a turn-level encoding of the therapeutic working alliance state at a turn-level resolution. We demonstrate on a real-world psychotherapy dialogue dataset that using this additional granular representation of the interaction dynamics between patients and therapists is beneficial both for interpretable post-session insights and diagnosing the patients from linguistic features. 

Although we concentrate on the Working Alliance Inventory in this article, our methodology is general and may be applied to a wider range of clinical assessment instruments. Ongoing next steps include deployment of our methods in clinical setting (e.g. using digital traces of medical records stored in real-time \cite{lin2022knowledge}), using topic modeling (as in \cite{lin2022neural}) to provide another interpretable dimension to boost our classification evaluation, incorporating additional predictive units of psychological states (as in \cite{lin2022predicting,lin2022predicting2})
and potentially training neuroscience-inspired reinforcement learning agents (as in \cite{lin2020story,lin2019split,lin2021models,lin2020unified}) to recommend best treatment strategy (as in \cite{lin2022supervisor}) using the classification labels informed by working alliance information as decision making contexts.





\bibliographystyle{IEEEbib}
\bibliography{main}

\end{document}